\documentclass[conference]{IEEEtran}
\IEEEoverridecommandlockouts
\usepackage{cite}
\usepackage{amsmath,amssymb,amsfonts}
\usepackage{algorithmic}
\usepackage{graphicx}
\graphicspath{{Figures/}}
\usepackage{textcomp}
\usepackage{xcolor}
\usepackage{svg}
\usepackage{caption}
\usepackage{subcaption}
\def\BibTeX{{\rm B\kern-.05em{\sc i\kern-.025em b}\kern-.08em
    T\kern-.1667em\lower.7ex\hbox{E}\kern-.125emX}}
\makeatletter
\newcommand{\linebreakand}{%
  \end{@IEEEauthorhalign}
  \hfill\mbox{}\par
  \mbox{}\hfill\begin{@IEEEauthorhalign}
}
\makeatother

\begin{document}

\title{Spike-time encoding of gas concentrations using neuromorphic analog sensory front-end}

\author{\IEEEauthorblockN{Shavika Rastogi\IEEEauthorrefmark{2}}
\IEEEauthorblockA{\textit{International Centre for Neuromorphic Systems} \\
\textit{Western Sydney University}\\
Sydney, Australia}
\and
\IEEEauthorblockN{Nik Dennler\IEEEauthorrefmark{3}}
\IEEEauthorblockA{\textit{Biocomputation Research Group} \\
\textit{University of Hertfordshire}\\
Hatfield, United Kingdom}
\and
\IEEEauthorblockN{Michael Schmuker}
\IEEEauthorblockA{\textit{Biocomputation Research Group} \\
\textit{University of Hertfordshire}\\
Hatfield, United Kingdom}
\linebreakand 
\IEEEauthorblockN{André van Schaik\IEEEauthorrefmark{1}}
\IEEEauthorblockA{\textit{International Centre for Neuromorphic Systems} \\
\textit{Western Sydney University}\\
Sydney, Australia}
\IEEEcompsocitemizethanks{\IEEEcompsocthanksitem\IEEEauthorrefmark{1}Corresponding author. Email: a.vanschaik@westernsydney.edu.au.}
\IEEEcompsocitemizethanks{\IEEEcompsocthanksitem\IEEEauthorrefmark{2}Also with Biocomputation Research Group, University of Hertfordshire.}
\IEEEcompsocitemizethanks{\IEEEcompsocthanksitem\IEEEauthorrefmark{3}Also with International Centre for Neuromorphic Systems, Western Sydney University.}
}

\maketitle

\begin{abstract}
Gas concentration detection is important for applications such as gas leakage monitoring. Metal Oxide (MOx) sensors show high sensitivities for specific gases, which makes them particularly useful for such monitoring applications. However, how to efficiently sample and further process the sensor responses remains an open question.
Here we propose a simple analog circuit design inspired by the spiking output of the mammalian olfactory bulb and by event-based vision sensors. Our circuit encodes the gas concentration in the time difference between the pulses of two separate pathways. We show that in the setting of controlled airflow-embedded gas injections, 
the time difference between the two generated pulses 
varies inversely with gas concentration, which is in agreement with the spike timing difference between tufted cells and mitral cells of the mammalian olfactory bulb. Encoding concentration information in analog spike timings may pave the way for rapid and efficient gas detection, and ultimately lead to data- and power-efficient monitoring devices to be deployed in uncontrolled and turbulent environments.
\end{abstract}

\begin{IEEEkeywords}
Metal Oxide Sensors, Olfactory Bulb, Gas concentration detection, Spike time encoding
\end{IEEEkeywords}

\section{Introduction}
Detecting gas is crucial in various areas around us to prevent health hazards. 
Apart from the identification of gases, it is often relevant to determine the concentration of a particular gas for safety reasons.  
For example, carbon monoxide is an odorless gas that can be released by gas and wood stoves due to incomplete combustion. Severe exposure can lead to brain and heart damage, or even death. Thus, there exists a demand for rapid concentration detection systems that can process gas sensor data rapidly and efficiently.

\par
In Neuromorphic Engineering, a primary objective is to devise algorithms that draw inspiration from the brain and enable the extraction of meaningful information from the external world \cite{Mead1990, Lyon1988}. Particularly, event-based sampling is an effective way of dealing with infrequent or intermittent, but rapidly changing stimuli, by maintaining a high temporal resolution when needed but reducing redundant information in the recorded data, which in return reduces storage requirements.

The emerging field of Neuromorphic Olfaction studies computational principles of biological olfactory systems, and translates them into algorithms and devices \cite{Chicca2014}. Some of the neuromorphic methods in the field of artificial olfaction try to emulate the entire biological olfactory pathway \cite{Koickal2007, Beyeler2010, Imam2012, Rivera2007}. 
Other methods focus on increasing the performance of gas sensing systems and targeted either towards modular developments such as gas sensing front ends \cite{Covington2003, Marco2014, Huang2017} and processing units \cite{Tang2011, Kasap2013, Martinelli2009, Hoda2010}, or towards the emulation of neurobiological principles that can be implemented in silicon \cite{Pan2012, Hung2012, Vanarse2017, Hausler:2011}. \par


Huang et al.\ \cite{Huang2017} developed a gas sensing front end that encoded the output of Metal Oxide (MOx) gas sensors \cite{Wang2010} for different gases into concentration invariant spike patterns. A feature extraction algorithm was used, in which the sensor response corresponding to each analyte was mapped to a single unique trajectory by logarithmic transformation. Others have focused on converting gas sensor data to spike trains, aiming to increase gas discrimination performance \cite{Martinelli2011, Yamani2012}, or to study the efficiency of an event-based approach \cite{dennler2022NICE}.
Han et al. \cite{Han2022} proposed an artificial olfactory neuron module comprising a chemo-resistive gas sensor and a one-transistor neuron, where the gas concentration was encoded in the spiking frequency of neurons. However, encoding concentration information in spike timings instead of spike frequencies is more efficient because the information is available after a single spike interval, as opposed to needing to average over many intervals for a rate code\cite{Rullen2001, Sengupta2017}. \par

In this paper, we propose an analog circuit design for spike time encoding of the concentration of a known gas. The circuit is inspired by the ATIS pixel design (Asynchronous Time-based Image Sensor) \cite{Posch2014, Posch2010} in event-based cameras. The concentration encoding mechanism of this circuit shares similarities with that of mitral cells and tufted cells - the two principal output neurons of mammalian olfactory bulb \cite{Fukunaga2012}. To our knowledge, this is the first attempt for encoding gas concentration levels measured with gas sensors in spike timings. \par

The paper is organized as follows: Section II describes the experimental setup used to obtain the recordings from the MOx sensor at different gas concentrations. Section III presents the proposed analog circuit design and describes the analogy it shares with the spiking output of the mammalian olfactory bulb and the ATIS pixel mechanism. Section IV presents the results obtained from circuit simulation for different concentrations of different gases. Finally, section V concludes the paper with discussions and planned future directions.

\section{Experimental Setup for MOx Sensor Recordings}

Electronic nose (e-nose) MOx sensor data for different gas concentrations were used (available online \cite{dennler_nik_2023_8299895}). The custom-made e-nose used to collect the data---similar to \cite{Drix2022RapidRO}---comprises four different Metal Oxide (MOx) gas sensors operated at a constant heater voltage, where our experiments use one of these sensors (the reducing sensor in the MiCS 6814). 
The odor stimuli were provided by a 
multi-channel odor delivery device that is described in detail by Ackels et al. \cite{Ackels2021}. It offers an exceptionally high temporal fidelity, which is achieved by combining high-speed gas valves, flow controllers, as well as short and narrow gas pathways. 
Odorant headspace samples were embedded in constant airflow and presented to the e-nose. \textit{EB}, \textit{Eu} and \textit{IA} gases were diluted in mineral oil at a ratio of $1:5$, and \textit{2H} was diluted in a ratio of 1:50. The delivered odor concentration was varied by modulating the valve shutter duty cycle, ensuring a linear relationship between different concentration levels. Fluctuations in flow during gas presentations were minimized by careful calibration.
For our experiments, recordings of four different odor stimuli were used: Ethyl Butyrate (\textit{EB}), Eucalyptol (\textit{Eu}), Isoamyl Acetate (\textit{IA}), and 2 - Heptanone (\textit{2H}).

\begin{figure}[h]
\includegraphics[width=\linewidth]{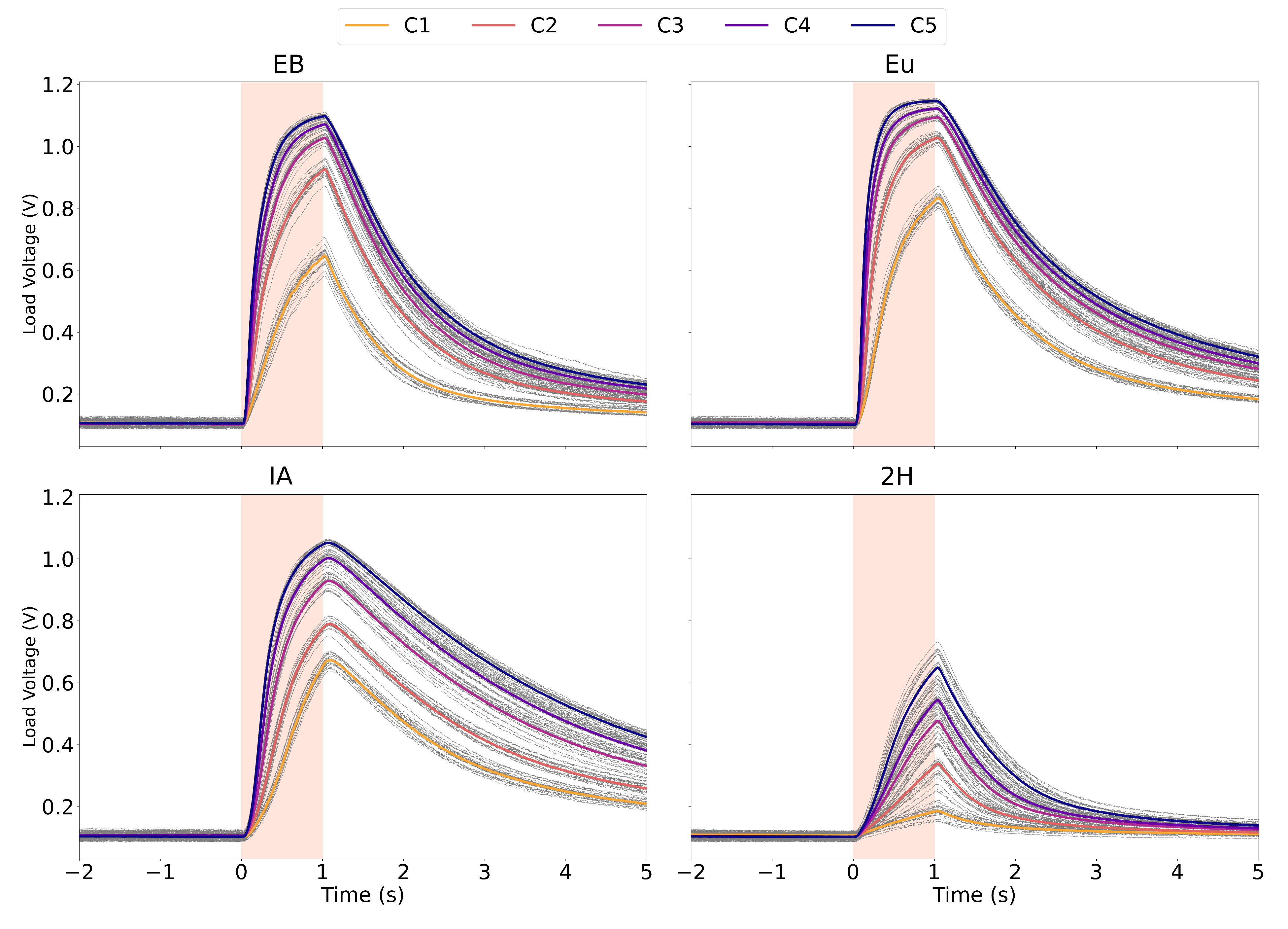}
\caption{Mean load voltage over all trials obtained for the sensor for four different gases at five different concentration levels. Gas stimulus is indicated by the shaded region, with the gas type indicated in the title of each panel. C1 to C5 indicate 5 concentration levels of each gas such that C1 is the lowest and C5 is the highest concentration level, and each colored curve indicates the mean response at these concentrations. The individual trials are shown in grey.}
\label{fig:1_conc_graph}
\end{figure}

The measurements consist of voltage recordings across the load resistor connected to the MOx sensors. This voltage varied inversely with respect to the MOx sensor resistance and was used as the input signal to the circuit described in the next section. 
The mean load voltage recordings over all trials for the sensor are shown in Figure \ref{fig:1_conc_graph}) (Load Resistance = $27 k\Omega$). Negative timestamps indicate the baseline response of the sensor before gas release. 

At $t=0 s$, the gas is released and continued for $1 s$ (indicated by the shaded region). Gas release stopped at $t=1 s$  and sensors were allowed to return to baseline for 30 s before the subsequent consecutive trial of the experiment started. Each experiment was repeated 20 times, where the order of odors and concentrations was randomized. 

\section{Proposed Analog Circuit Design}
\subsection{Concentration Encoding in Mammalian Olfactory Bulb}
\begin{figure}
   \includegraphics[width=\linewidth]{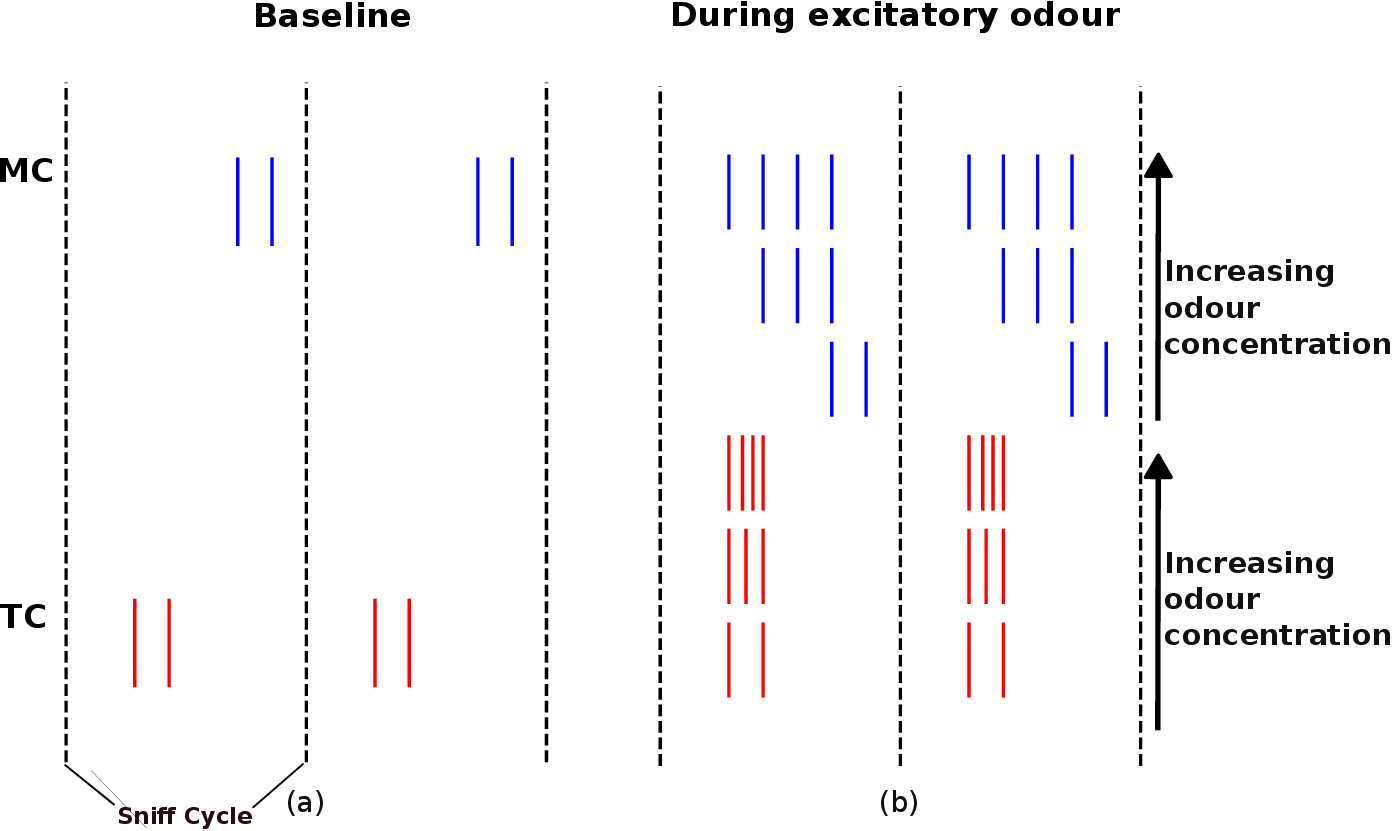}
   \caption{(a) Baseline firing of MCs and TCs. Both cells fire in opposite phases of the sniff cycle. (b) Variation in MC and TC firing with increasing odor concentration, as indicated by the arrows. Figures adapted from \cite{Fukunaga2012}.}
\label{fig:2_tc_mc}
\end{figure}

The Principal Neurons of the mammalian olfactory bulb, Mitral cells (MC) and Tufted cells (TC), fire in distinct and opposite phases of a sniff cycle in response to an odor stimulus as shown in Figure \ref{fig:2_tc_mc}(a) \cite{Fukunaga2012}, with TCs firing earlier than MCs. Weak excitation and strong inhibition are factors responsible for delayed MC firing \cite{Fukunaga2012, Geramita2017}. As odor concentration increases, MCs spike earlier in the sniff cycle. TCs respond to concentration increases with a firing rate increase at a constant average phase in the sniff cycle \cite{Fukunaga2012} (Figure \ref{fig:2_tc_mc}(b)). Thus, odor concentration in the olfactory bulb is encoded in the time difference between MC and TC firing.

\subsection{ATIS Pixel Mechanism}\label{AA}
The Asynchronous Time-based Image Sensor (ATIS) is a bioinspired image sensor driven by changes in a visual scene. The ATIS combines the advantages of event-driven data acquisition and time domain spike encoding of image information \cite{Posch2014}.
In an ATIS pixel, a change detection circuit generates an event when a significant change in light intensity is detected. This event in turn activates an exposure measurement circuit that generates two events with a delay between them that is inversely proportional to the light intensity \cite{Posch2010}, similar to the delay between spikes from MCs and TCs in the olfactory bulb.

\subsection{Proposed Circuit Design}
\begin{figure}[h]
\includegraphics[width=\linewidth]{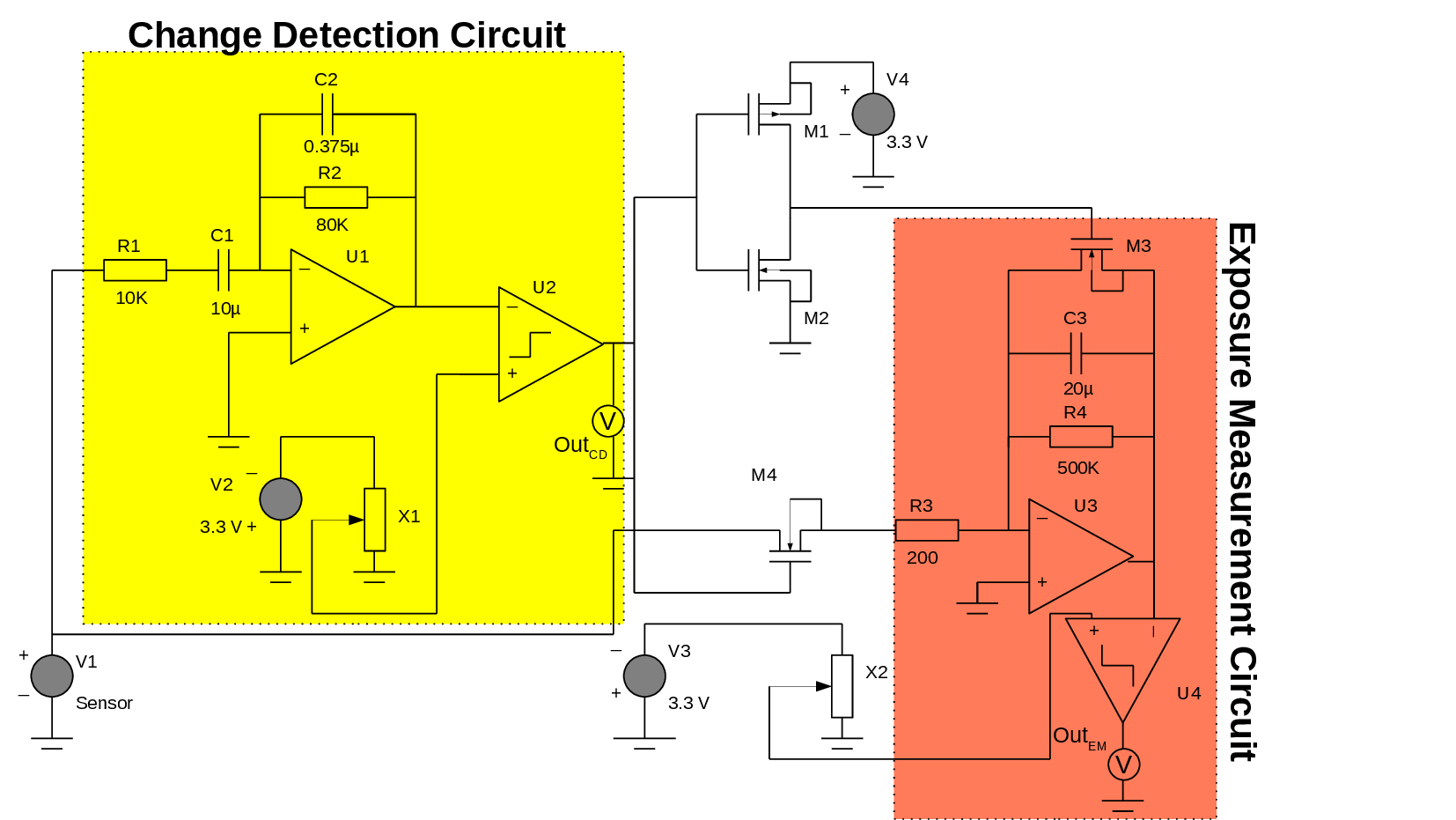}
\caption{Proposed gas concentration measurement circuit.}
\label{fig:3_circuit}
\end{figure}

Taking inspiration from the ATIS pixel circuitry which encodes light intensity information in the timings of two consecutive events, we designed a circuit for encoding gas concentration levels in analog spike timings (see Figure \ref{fig:3_circuit}). 
Similar to the ATIS pixel circuit, it consists of two sub-circuits: 1. A Change detection circuit, 2. An exposure measurement circuit. The MOx sensor is connected in series with a load resistor, across which the voltage drop V1 serves as an input to both subcircuits.

The change detection (CD) circuit 
comprises an inverting differentiator followed by an op-amp comparator. The inverting differentiator output is compared with a pre-defined threshold voltage by the comparator. A change detection pulse ($Out_{CD}$) is generated whenever the differentiator output exceeds a threshold, 
which is adjustable by the potentiometer X1 and can be varied depending on the SNR of the input signal. 

The exposure measurement (EM) circuit takes input from the transistor M4, which is switched on by the change detection pulse.
This subcircuit comprises an op-amp integrator followed by an op-amp comparator. The integrator output is compared to the threshold set by potentiometer X2. 
An exposure measurement pulse ($Out_{EM}$) is generated whenever the integrator output exceeds this threshold value. The falling edge of the change detection pulse activates a CMOS inverter (comprising of PMOS transistor M1 and NMOS transistor M2). This 
activates transistor M3, which resets the integrator. The exposure measurement stops when the integrator output falls below the threshold. 

The timing of the rising flanks of the change detection pulse and the exposure measurement pulse respectively are recorded. Their time difference may encode an estimate of the concentration of a particular gas.\par

The circuit parameters were tuned on one trial of each gas at different concentration levels, and tested on the remaining trials. For the differentiator in the CD circuit, the time constants are chosen such that it has a high gain during the rising flank of the input signal (when the gas is released), and it reverts as soon as the input signal starts falling (when the gas release is stopped). For the integrator in the EM circuit, the feedback resistor and capacitor values are carefully selected to prevent the integrator response from saturating within the given range of odor concentrations. Additionally, these values ensure that the time constant remains within the desired range. 
We used \texttt{Altium Designer} \cite{altium} for the design and simulation of the circuit.

\section{Results}
\begin{figure}[h]
  \includegraphics[width=\linewidth]{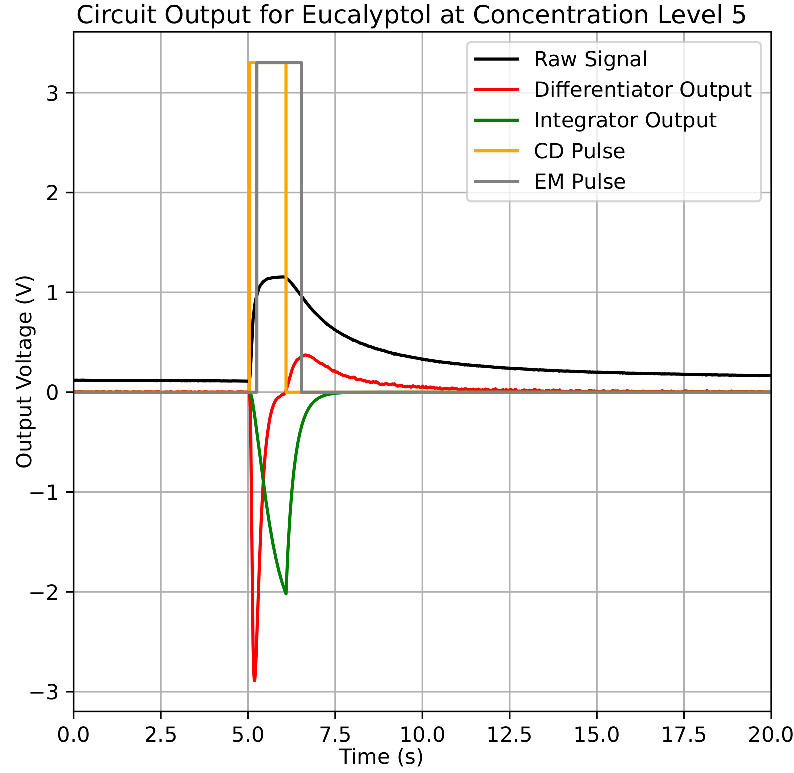}
    \caption{Output of circuit for Eucalyptol (Eu) gas at concentration level 5}
\label{fig:4_circuit_output}
\end{figure}

Figure \ref{fig:4_circuit_output} displays the circuit simulation results for one trial using Eucalyptol (Eu) gas at concentration level 5. From the inverting differentiator output, it can be seen that the response increases (in a negative direction because of its inverting nature) during the rising flank of the signal and switches sign when the input signal starts decreasing. In this way, the CD pulse is on only for the rising flank of the signal. The response of the inverting integrator starts falling as soon as the CD pulse is off. The EM pulse takes some time to reset after the CD pulse is off. This is due to the higher decay time constant of the integrator.\par

\begin{figure}[h]
  \includegraphics[width=\linewidth]{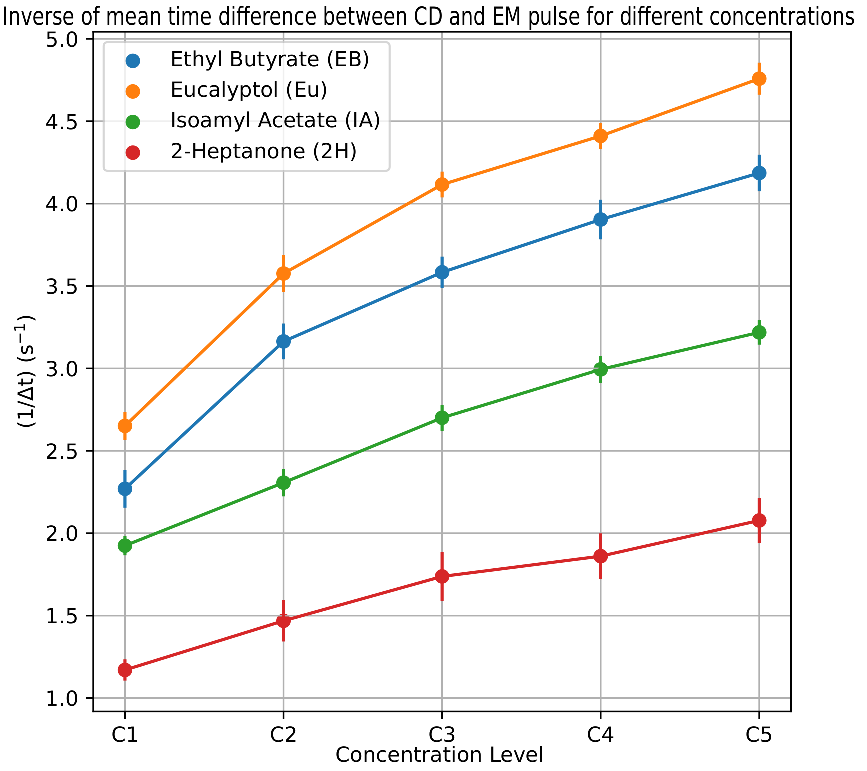}
    \caption{Inverse of the time difference between CD and EM pulse activation with respect to concentration level. Dots and error bars represent the mean and standard deviation across 20 trials respectively. For one trial for 2H at C1, the EM pulse was not observed and thus the trial was discarded.
    }
\label{fig:5_dt_vs_conc}
\end{figure}

We measured the difference in timings of activation of the CD and EM pulses for all gases at all concentration levels for this circuit. Figure \ref{fig:5_dt_vs_conc} shows the plot of the inverse of the mean time difference (over all trials) between CD and EM pulse activation with respect to the concentration level of each gas. Error bars at each concentration level represent the standard deviation over all trials. It can be observed from the graphs that the time difference between CD and EM pulses varies inversely with gas concentration. The rate of variation is not the same for all gases because a particular MOx sensor has different sensitivity for different gases. Thus, only the concentration level of a known gas can be decoded through this circuit by measuring the time difference between CD and EM pulses. For converting these finite-duration analog pulses into short spikes as observed in biology, or as used in the ATIS sensor, we can use an edge detector circuit (e.g.: a differentiator) to generate spikes on the positive edge of the CD and EM pulses. \par

\section{Discussion and Future Work}
The present work demonstrates a simple analog circuit implementation inspired by the ATIS pixel mechanism and the mammalian olfactory bulb for spike time encoding of gas concentrations. This circuit can be used in a  scenario where the gas sensor is exposed to a single gas pulse of varying concentrations. The circuit generates output only when there is some significant change in the environment due to gas injection, which is in contrast to ADCs which sample the gas sensor data even though there is no useful information. Therefore, this circuit is an efficient substitute for ADCs that can be interfaced with the MOx sensors.\par

One limitation of MOx sensors is drift, where the sensor response to the same gas changes over time, especially over longer intervals like days or months \cite{Dennler2022}. 
We observed little trial-to-trial variability in the data used here, which has been collected over 12 hours (see the error bars in Figure \ref{fig:5_dt_vs_conc}). Considering longer-term experiments, we expect the circuit to be resilient against DC resistance baseline drift due to its operation as a differentiator. The usage of two separate signal processing pathways (differentiation and integration) and the computation of the difference between their respective pulse timings could add additional resilience to higher-order drift. Additional longer-term exposure experiments will be required to fully assess how drift affects the circuit.\par
In the future, it will be of interest to extend the operation of the circuit to multiple sensors. Further, it has been shown that event-based analysis of concentration fluctuations encodes information about the odor source location \cite{Schmuker2016}. The event-based sensing concept could be applied in situations where instantaneous gas concentration changes rapidly, like in turbulent environments and in the context of mobile robotic olfaction \cite{Vuka2017}. If successful, this circuit could find use in many applications, such as advanced gas detectors in homes and industries e.g.\ for rapid gas leakage detection and air quality monitoring.


\section*{Acknowledgment}

Part of this work was funded by an NSF/MRC award under the Next Generation Networks for Neuroscience initiative (NeuroNex Odor to action, NSF \#2014217, MRC \#MR/T046759/1). 

\bibliographystyle{ieeetr}
\bibliography{biocas_ref}

\vspace{12pt}
\color{red}
\end{document}